# Generating Graphoids from Generalised Conditional Probability

Nic Wilson
Department of Computer Science
Queen Mary and Westfield College
Mile End Rd., London E1 4NS, UK
nic@dcs.qmw.uk.ac

## Abstract

We take a general approach to uncertainty on product spaces, and give sufficient conditions for the independence structures of uncertainty measures to satisfy graphoid properties. Since these conditions are arguably more intuitive than some of the graphoid properties, they can be viewed as explanations why probability and certain other formalisms generate graphoids. The conditions include a sufficient condition for the Intersection property which can still apply even if there is a strong logical relationship between the variables. We indicate how these results can be used to produce theories of qualitative conditional probability which are semi-graphoids and graphoids.

## 1 INTRODUCTION

The importance of the qualitative features of probabilistic reasoning has often been emphasised in the recent AI literature, especially by Judea Pearl. An important qualitative aspect of probability is given by the graphoid properties, defined in [Pearl, 88] (see also [Dawid, 79; Smith, 90]) which sum up many of the properties of probabilistic conditional independence. In this paper we look at the reasons why probability obeys these properties, with an eye to generating other uncertainty theories which share much of the same structure as probability, but represent different types of information, perhaps of a more qualitative nature.

A fairly general family of uncertainty calculi on product spaces is introduced, which we call *Generalised Conditional Probability on Product Spaces (GCPP)*, and define two different types of conditional independence for GCPPs. We show that under simple (and apparently fairly weak) conditions, conditional independence for GCPPs satisfies the graphoid properties. This means that then independence assumptions can be propagated using the graphoid inference rules, and can be represented and propagated using the (both directed and undirected) graphical methods of [Pearl, 88].

In section 2 we define independence structures, semi-graphoids and graphoids. GCPPs are defined in section 3, with examples and we show how they give rise to independence structures. Section 4 considers the Intersection property. In the literature it seems to be generally assumed that this only holds for probability distributions which are always non-zero; we show here that it holds much more generally, a sufficient condition being a connectivity property on the non-zero values of the probability; exactly the same condition is sufficient for other GCPPs to satisfy Intersection. Section 5 considers different sufficient conditions for GCPPs to give rise to semi-graphoids. These are useful for constructing uncertainty calculi which generate graphoids and might also be used for showing that an uncertainty calculus gives rise to a graphoid.

In section 6 we consider another view of GCPPs: as qualitative conditional probabilities. This view allows graphoids to be constructed from qualitative comparative judgements of probability. Section 7 briefly considers computation of GCPPs, and section 8 highlights some areas for further study.

## 2 INDEPENDENCE STRUCTURES

Let $U$ be a finite set. An independence structure $I$ on $U$ is defined to be a set of triples $(X, Z, Y)$ where $X, Y$ and $Z$ are disjoint[1] subsets of $U$. We write $I(X, Z, Y)$ for $(X, Z, Y) \in I$. For disjoint subsets $X, Y \subseteq U$, their union $X \cup Y$ will be written $XY$.

$U$ is intended to be a set of variables, and $I(X, Z, Y)$ is intended to mean that variables $X$ are independent of variables $Y$ given we know the values of the variables $Z$.

---

[1] A collection $\mathcal{A}$ of sets is said to be disjoint if for each $X, Y \in \mathcal{A}$, $X \cap Y = \emptyset$.

**The Graphoid Properties of Independence Structures**

$I(X, Z, \emptyset)$  (Trivial Independence)

If $I(X, Z, Y)$ then $I(Y, Z, X)$  (Symmetry)

If $I(X, Z, YW)$ then $I(X, Z, Y)$  (Decomposition)

If $I(X, Z, YW)$ then $I(X, ZY, W)$  (Weak Union)

If $I(X, ZY, W)$ and $I(X, Z, Y)$ then $I(X, Z, YW)$ (Contraction)

If $I(X, ZY, W)$ and $I(X, ZW, Y)$ then $I(X, Z, YW)$ (Intersection)

where $W, X, Y, Z$ are arbitrary disjoint subsets of $U$ (so, for example, $I$ satisfies symmetry if and only if the above property holds for all disjoint $X$, $Y$ and $Z$).

If an independence structure satisfies all these properties then it is said to be a graphoid; if it satisfies the first five (i.e, all except Intersection) then it is said to be a semi-graphoid. As we shall see in section 5, probabilistic conditional independence is a semi-graphoid, and in certain situations a graphoid.

The definitions given here for semi-graphoid and graphoid differ from that given in [Pearl, 88], in that we require Trivial Independence to hold. However, our definition seems to be what Pearl intended[2]; it is not implied by other properties (consider the empty independence structure) and it is satisfied by probabilistic conditional independence so it is a (rather trivial) counter-example to the Completeness Conjecture[3] [Pearl, 88]; also without Trivial Independence, Markov boundaries don't necessarily exist (consider the empty independence structure again) which makes Theorem 4 of [Pearl, 88] incorrect.

The intersection of a family of (semi-)graphoids is a (semi-)graphoid. Hence, for any independence structure $I$, there is a unique smallest (semi-)graphoid containing $I$.

## 3 GENERALISED CONDITIONAL PROBABILITY ON PRODUCT SPACES (GCPPs)

Uncertainty measures are usually defined on boolean algebras. However, for our purposes of studying independence structures generated by the uncertainty measure, a different domain is natural.

### 3.1 THE BASIC DEFINITIONS

$U = \{X_1, \ldots, X_n\}$ is said to be a set of variables if associated with each variable $X_i \in U$ is a finite set of values $\underline{X_i}$. For $X \subseteq U$ define $\underline{X}$ to be $\prod_{X_i \in X} \underline{X_i}$. For disjoint $X, Y \subseteq U$, an element of $\underline{XY}$ may be written $xy$ for $x \in \underline{X}$, $y \in \underline{Y}$. For disjoint $X, Y \subseteq U$ we define $\underline{X|Y}$ to be the set of all pairs $\{x|y : x \in \underline{X}, y \in \underline{Y}\}$. The set $\underline{\emptyset}$ is defined to be a singleton $\{\top\}$. An element $x|\top$ of $\underline{X}|\underline{\emptyset}$ will usually be abbreviated to $x$ (we are identifying $\underline{X}|\underline{\emptyset}$ with $\underline{X}$).

---

[2] See, for example, the sentence before Theorem 4, p97 of [Pearl, 88].

[3] Fatal counter-examples are given in [Studený, 92].

The set $U^*$ is defined to be $\bigcup_{\substack{X,Y \subseteq U \\ X \cap Y = \emptyset}} \underline{X|Y}$.

A GCPP $\rho$ over set of variables $U$ is defined to be a function $\rho: U^* \to D$ for some set $D$ containing different distinguished elements 0 and $\infty$ such that for any disjoint $X, Y \subseteq U$ and $x \in \underline{X}$,

(i) $\rho(x) = 0$ if and only if for all $y \in \underline{Y}$, $\rho(xy) = 0$, and

(ii) for any $y \in \underline{Y}$, $\rho(x|y) = \infty$ if and only if $\rho(y) = 0$.

GCPPs will be viewed as measures of uncertainty; $\rho(x|y)$ may be thought of as some sort of measure of how plausible it is that the composite variable $X$ takes the value $x$, given that $Y$ takes the value $y$. The assignment $\rho(x|y) = 0$ is intended to mean that $x$ is impossible given $y$, i.e., $X$ cannot take the value $x$ if $Y$ takes the value $y$. The inclusion of element $\infty$ in $D$ is not strictly necessary; it is used as a notational convenience, and can be read as 'undefined'. We require (i) because: $x$ is possible if and only if there is some value of $Y$ which is possible when $X$ takes the value $x$. We require (ii) because: if $y$ is impossible then conditioning on $y$ doesn't make much sense.

Note that no structure on $D$ is assumed; for example, we do not assume that $D \setminus \{\infty\} \subseteq \mathbb{R}$, or even that $D$ has an ordering on it.

The definition implies that $\rho(\top) \neq 0, \infty$ and that for any $X \subseteq U$ and $x \in \underline{X}$, $\rho(x) \neq \infty$.

**Definition**

For GCPP $\rho$ over $U$ and disjoint $X, Y \subseteq U$ define $\rho^{X|Y}$ to be $\rho$ restricted to $\underline{X|Y}$ and define $\rho^X$ to be $\rho^{X|\emptyset}$. For $Z \subseteq U$ and $z \in \underline{Z}$ define $\rho_z: (U \setminus Z)^* \to D$ by, for disjoint $X, Y \subseteq U \setminus Z$, $x \in \underline{X}$, $y \in \underline{Y}$, $\rho_z(x|y) = \rho(x|yz)$. For disjoint $X, Y \subseteq U \setminus Z$, $\rho_z^{X|Y}$ is defined to be $\rho_z$ restricted to $\underline{X|Y}$, and $\rho_z^X$ is defined to be $\rho_z^{X|\emptyset}$.

GCPP $\rho$ over $U$ is said to be a full GCPP over $U$ if for every $Z \subseteq U$ and $z \in \underline{Z}$ such that $\rho(z) \neq 0$, $\rho_z$ is a GCPP over $\overline{U} \setminus Z$.

The function $\rho_z$ may be thought of as $\rho$ conditioned on $Z = z$. It turns out that, for GCPP $\rho$, $\rho$ is a full GCPP if and only if for all disjoint $X, Y \subseteq U$, $x \in \underline{X}$, $y \in \underline{Y}$, $[\rho(x|y) = 0 \iff \rho(xy) = 0 \text{ and } \rho(y) \neq 0]$.

### 3.2 EXAMPLES OF GCPPS

A probability function over set of variables $U$ is defined to be a function $P: U^* \to [0, 1] \cup \{\infty\}$ such that for $x|y \in U^*$, (i) $P(x|y) = \infty \iff P(y) = 0$; (ii) if $P(y) \neq 0$, $P(x|y) = P(xy)/P(y)$; and (iii) $P(x) = \sum_{w \in \underline{U \setminus X}} P(xw)$.

The definition implies that P is a full GCPP over $U$ and $P(\top) = 1$. The latter follows since $P(\top)$ is equal, by definition, to $P(\top|\top)$ so by (i) above, $P(\top) = 0$ if and only if $P(\top) = \infty$, which implies that $P(\top)$ is neither 0 or $\infty$. We can now apply (ii) to get $P(\top) = P(\top|\top) = P(\top)/P(\top)$ which implies that $P(\top) = 1$ as required.

For any (finite) set of variables $U$, there is a one-to-one correspondence between probability functions P over $U$ and probability functions $f$ on $\underline{U}$, i.e., functions $f:\underline{U} \to [0,1]$ such that $\sum_{u \in \underline{U}} f(u) = 1$; P restricted to $\underline{U}$ is a probability function on $\underline{U}$, and conversely, a probability function $f$ on $\underline{U}$ extends uniquely to a probability function over $U$ using (i), (ii) and (iii).

A Dempster possibility function over set of variables $U$ is defined to be a function $\pi: U^* \to [0,1] \cup \{\infty\}$ such that for $x|y \in U^*$, (i) $\pi(x|y) = \infty \iff \pi(y) = 0$; (ii) if $\pi(y) \neq 0$, $\pi(x|y) = \pi(xy)/\pi(y)$; and (iii) $\pi(x) = \max_{w \in \underline{U \setminus X}} \pi(xw)$.

Again, the definition implies that $\pi$ is a full GCPP over $U$ and $\pi(\top) = 1$. Dempster possibility functions are essentially Zadeh's possibility measures [Zadeh,78; Dubois and Prade, 88] and consonant plausibility functions [Shafer, 76]; the definition of conditional possibility is obtained from Dempster's rule, and is not the one most commonly used, partly because it means that the range of $\pi$ cannot be viewed as an ordinal scale (and most justifications of possibility theory require this).

A special case of Dempster possibility functions are *consistency functions over* $U$, where $\pi$ only takes the values 0, 1 and $\infty$. $\pi(x|y) = 1$ is then intended to mean that, given that $y$ is the true value of variables $Y$, it is possible that $x$ is the true value of variables $X$. Every full GCPP $\rho$ over $U$ gives rise to a consistency function $\rho^*$ over $U$ defined, for $\psi \in U^*$, by $\rho^*(\psi) = \rho(\psi)$ if $\rho(\psi) = 0$ or $\infty$, and $\rho^*(\psi) = 1$ otherwise. Consistency functions appear in the theory of relational databases [Fagin, 77], and also in [Shafer et al., 87].

A kappa function over set of variables $U$ is defined to be a function $\kappa: U^* \to \{0, 1, 2, \ldots, \infty, \infty_D\}$ (where $\infty_D$ is different from the other elements) such that for $x|y \in U^*$, (i) $\kappa(x|y) = \infty_D \iff \kappa(y) = \infty$; (ii) if $\kappa(y) \neq \infty$, $\kappa(x|y) = \kappa(xy) - \kappa(y)$; and (iii) $\kappa(x) = \min_{w \in \underline{U \setminus X}} \kappa(xw)$.

The definition implies that $\kappa(\top) = 0$ and $\kappa$ is a full GCPP over $U$ (however, the labelling of the elements in the range of $\kappa$ is confusing: the zero of $D$ in the definition of a GCPP is $\infty$ and the element meaning 'undefined' is $\infty_D$ not $\infty$). Kappa functions are based on Spohn's Ordinal Conditional Functions [Spohn, 88]. An important difference is that the range of kappa functions has a maximum element $\infty$; Spohn did not include a maximal element in the range of OCFs because he desired belief change to be reversible [Spohn, 88, p130].

Kappa function $\kappa$ can be transformed into a Dempster possibility function $\pi^\kappa$ by $\pi^\kappa(\psi) = \infty \iff \kappa(\psi) = \infty_D$ and $\pi^\kappa(\psi) = 2^{-\kappa(\psi)}$ otherwise, for $\psi \in U^*$, where $2^{-\infty}$ is taken to be 0. For $\psi, \phi \in U^*$, $\kappa(\psi) = \kappa(\phi) \iff \pi^\kappa(\psi) = \pi^\kappa(\phi)$. This means that, for our purposes, kappa functions can be viewed as special cases of Dempster possibility functions.

Shafer's plausibility functions [Shafer, 76], also give full GCPPs; their dual functions, belief functions, and necessity functions, the dual of possibility functions, do not give GCPPs, since, for these, a value of 0 means a lack of evidence, rather than 'impossible'.

### 3.3 INDEPENDENCE STRUCTURES OF GCPPS

For GCPP $\rho$ over set of variables $U$, independence structures $I_\rho$ and $I'_\rho$ are defined as follows. Let $X$, $Y$ and $Z$ be disjoint subsets of $U$.

$I_\rho(X, Z, Y)$ if and only if $\rho(x|yz) = \rho(x|z)$ for all $x \in \underline{X}$, $y \in \underline{Y}$, $z \in \underline{Z}$ such that $\rho(yz) \neq 0$.

$I'_\rho(X, Z, Y)$ if and only if $\rho(x|yz) = \rho(x|y'z)$ for all $x \in \underline{X}$, $y, y' \in \underline{Y}$, $z \in \underline{Z}$ such that $\rho(yz) \neq 0$ and $\rho(y'z) \neq 0$.

For set $S$ and functions $g, h: S \to D$ write $g =^\infty h$ if they are equal when they are both defined, i.e., if for all $s \in S$, $[g(s) = h(s)$ or $g(s) = \infty$ or $h(s) = \infty]$. This gives a simpler way of expressing the two independence structures. For disjoint subsets $X$, $Y$ and $Z$ of $U$,

$I_\rho(X, Z, Y)$ if and only if for all $y \in \underline{Y}$, $\rho_y^{X|Z} =^\infty \rho^{X|Z}$, and

$I'_\rho(X, Z, Y)$ if and only if for all $y, y' \in \underline{Y}$,

$\rho_y^{X|Z} =^\infty \rho_{y'}^{X|Z}$.

To understand the definitions, first consider the case when $Z = \emptyset$. Then $I_\rho(X, Z, Y)$ if the degree of plausibility of $x$, $\rho(x)$, does not change when we condition by any (possible) value $y$ of $Y$. Thus our uncertainty about variable $X$ does not change by learning the value of variable $Y$. $I'_\rho(X, Z, Y)$ holds if the degree of plausibility of $x$ given $y$, $\rho(x|y)$ does not depend on the choice of (possible) value $y$ of $Y$. The same remarks apply for general $Z$, except that now we must consider the degrees of plausibility conditional on each value $z$ of $Z$.

We shall see in section 5 that for any GCPP $\rho$, $I_\rho$ satisfies Trivial Independence and Contraction, and, if $I_\rho = I'_\rho$, it satisfies Decomposition and Weak Union also.

## 4  THE INTERSECTION PROPERTY

This is the only one of the graphoid properties that does not hold for all probability functions. [Pearl, 88, p87] appears to suggest that it only holds for probability functions which are everywhere non-zero. This turns out not to be the case, and we will see that a sufficient condition for Intersection to hold sometimes allows very strong logical dependencies between the variables.

Set $\Omega$ is said to be connected under relation $R \subseteq \Omega \times \Omega$ if the smallest equivalence relation on $\Omega$ containing $R$ is the relation $\Omega \times \Omega$.

Let $\rho$ be a GCPP over set of variables $U$ and let $Y$, $W$ and $Z$ be disjoint subsets of $U$. For $z \in \underline{Z}$ define $(\underline{YW})^+_{\rho,z} = \{yw \in \underline{YW} : \rho(ywz) \neq 0\}$. We say that $(Y, W)$ is $\rho, Z$-connected[4] if for all $z \in \underline{Z}$, $(\underline{YW})^+_{\rho,z}$ is connected under the relation $R$ defined by $yw\, R\, y'w' \iff y = y'$ or $w = w'$.

For GCPP $\rho$ over set of variables $U$, we say that $U$ if $\rho$-connected if for all disjoint subsets $Y, W, Z$ of $U$, the pair $(Y, W)$ is $\rho, Z$-connected.

Note that these properties only depend on the set of elements of $\underline{U}$ for which $\rho$ is zero (that is, those which are known to be impossible).

The above concepts are not quite as obscure as they appear at first sight. $(\underline{YW})^+_{\rho,z}$ is the set of $yw$ which are not known to be impossible when we know that $Z = z$. If we label $\underline{Y}$ as $y_1, \ldots, y_m$ and $\underline{W}$ as $w_1, \ldots, w_n$ then $\underline{YW} = \underline{Y} \times \underline{W}$ can be viewed as the squares of a $m \times n$ chess board. Then $y_i w_j\, R\, y_{i'} w_{j'}$ iff $i = i'$ or $j = j'$, i.e., iff the rook chesspiece could move between the squares $(i, j)$ and $(i', j')$. Let $N_z$ be the set of squares corresponding to $(\underline{YW})^+_{\rho,z}$. We therefore have that $(Y, W)$ is $\rho, Z$-connected iff for all $z$, it is possible to move between any two elements of $N_z$ using a sequence of rook moves, where each intermediate square is also in $N_z$.

### Proposition 1

Let $\rho$ be a GCPP over a set of variables $U$.

(i) For disjoint subsets $X$, $Y$, $Z$ and $W$ of $U$, suppose that $(Y, W)$ is $\rho, Z$-connected and also that $I'_\rho(X, ZY, W)$ and $I'_\rho(X, ZW, Y)$. Then $I'_\rho(X, Z, YW)$ holds.

(ii) If GCPP $\rho$ over set of variables $U$ is such that $U$ if $\rho$-connected then $I'_\rho$ satisfies Intersection.[5]

---

[4] Interestingly, a very similar concept is important in Dempster-Shafer theory: in [Moral and Wilson, 94] it guarantees the convergence of Markov Chain Monte-Carlo algorithms, and in [Wilson, 93] it is relevant to the justification of Dempster's rule.

[5] Milan Studený [89] has found a similar result (for the case of probability functions).

If GCPP $\rho$ is non-zero on $\underline{U}$ then, trivially, $U$ is $\rho$-connected, and so $I'_\rho$ satisfies Intersection.

### Example

Let $U = \{S, H_1, H_2\}$. Variable $S$ ranges over shoe sizes, and the correct value is the shoe size of the (unknown) next person to walk into my office. $H_1$ and $H_2$ both take integer values between 0 and 3000. The correct value of $H_1$ is the height in millimetres rounded down of this unknown person and the correct value of $H_2$ is their height to the nearest millimetre.

Let P be a Bayesian probability function on $\underline{U}$, representing our Bayesian beliefs about the variables. As described above, P extends uniquely to a GCPP over $U$. Now, $P(ij) = 0$ unless $i = j$ or $i = j - 1$, where $ij$ means $H_1 = i$ and $H_2 = j$. Despite the very strong logical relationship between $H_1$ and $H_2$, $(\{H_1\}, \{H_2\})$ is $P, \emptyset$-connected, and so if we considered $S$ to be logically independent of $\{H_1, H_2\}$, in the sense that $P(sh_1h_2) = 0$ if and only if $P(s) = 0$ or $P(h_1h_2) = 0$, then $U$ would be P-connected. This implies that $I_P$ ($= I'_P$ by the results of the next section) would satisfy the Intersection axiom, and so would be a graphoid.

In any case, given knowledge of height to the nearest millimetre, one will learn almost nothing more about shoe size by learning height in millimetres rounded down, so one might be tempted to make the subjective conditional independence judgement $I_P(\{S\}, \{H_2\}, \{H_1\})$. (Alternatively, if one did not know the precise meaning of $H_1$ and $H_2$, but had the values of the variables for previous visitors then it would take a vast amount of data to detect a dependency.) Similarly one might be tempted to say $I_P(\{S\}, \{H_1\}, \{H_2\})$. But these lead, using the last proposition, to $I_P(\{S\}, \emptyset, \{H_1, H_2\})$ which is certainly unreasonable since there is a definite dependency between shoe size and height.

This example illustrates how careful one must be with subjective independence judgements for Bayesian probability (or any other GCPP). It also seems to suggest that GCPPs, with definition $I'_\rho$, cannot represent 'approximate independence'.

## 5  SUFFICIENT CONDITIONS FOR GCPPS TO GENERATE GRAPHOIDS

Here we consider simple sufficient conditions on GCPPs for its associated independence structures to satisfy semi-graphoid properties. Since probability functions, Dempster possibility functions, kappa functions and consistency functions satisfy these properties with the exception of the conditions of proposition 4(v), these could be viewed as explanations for why they are semi-graphoids.

## 5.1 CONDITIONAL COHERENCE

It is easy to see that for any GCPP $\rho$, $I_\rho \subseteq I'_\rho$, i.e., if $I_\rho(X, Z, Y)$ holds then $I'_\rho(X, Z, Y)$ must hold. Furthermore if $\rho$ is intended to represent a generalised form of probability then a natural constraint is that $I_\rho = I'_\rho$. This is because a sufficient condition for $I_\rho = I'_\rho$ is the apparently reasonable condition:

*Conditional-Coherence:* For any disjoint subsets $X, Y, Z$ of $U$, $x \in \underline{X}$ and $z \in \underline{Z}$, if for all $y, y' \in \underline{Y}$, $\rho(x|yz) = \rho(x|y'z)$ (i.e., $\rho(x|yz)$ does not vary with $y$) then $\rho(x|z) = \rho(x|yz)$ (i.e., $\rho(x|z)$ is equal to that constant value).

We say that $\rho$ is *weakly conditional-coherent* if $I_\rho = I'_\rho$.

Consider conditioning on a fixed $z \in \underline{Z}$. The idea behind conditional coherence is that if the degree of plausibility of $x$ given $y$ (i.e, $\rho_z(x|y)$) is not dependent on which value $y$ of $Y$ we use, then one might expect that the degree of plausibility of $x$ (i.e., $\rho_z(x)$) would be equal to that constant value. The conditionals and marginal then cohere in a particular sense.

Conditional-Coherence is a restricted version of the Sandwich Principle [Pearl, 90; IJAR, 92]. Plausibility/belief functions and upper/lower probability functions have good reason not to obey conditional coherence: see e.g., [Wilson, 92; Chunhai and Arasta, 94]. It is satisfied by probability and Dempster possibility functions and hence by consistency and kappa functions.

### Proposition 2

For any GCPP $\rho$ over set of variables $U$ the independence structure $I_\rho$ satisfies Trivial Independence and Contraction, and $I_\rho$ satisfies Decomposition if and only if it satisfies Weak Union.

Now suppose that $\rho$ is weakly conditional-coherent. We then have

(i) $I_\rho$ satisfies Decomposition and Weak Union. Therefore if $I_\rho$ satisfies Symmetry then it is a semi-graphoid.

(ii) If $U$ is $\rho$-connected then $I_\rho$ satisfies Intersection. Therefore if $I_\rho$ satisfies Symmetry then it is a graphoid.

Perhaps the only one of the graphoid properties, with the exception of Trivial Independence, which immediately seems natural is Symmetry. Surprisingly, it seems to be harder to find natural sufficient conditions on $\rho$ for $I_\rho$ to satisfy Symmetry. The following result gives a fairly strong condition.

### Proposition 3

Suppose that $\rho$ is a full GCPP over $U$ such that for all $Z \subseteq U$ and $z \in \underline{Z}$, there exists a function $\diamond: R \to D$ for some $R \subseteq D \times D$ such that

(i) for all $x|y \in (U \setminus Z)^*$, $\rho_z(xy) = \rho_z(x|y) \diamond \rho_z(y)$, and

(ii) if $a \diamond b = c \diamond a$ and $a \neq 0$ then $b = c$.

Then $I_\rho$ satisfies Symmetry.

## 5.2 DETERMINISTIC RELATIONS BETWEEN JOINTS, CONDITIONALS AND MARGINALS

If $I$ is an independence structure let its reflection $I^R$ be defined by $I^R(Y, Z, X) \iff I(X, Z, Y)$. Clearly, $I$ satisfies Symmetry if and only if $I = I^R$. Let $I^S = I \cap I^R$ be the symmetric part of $I$, so that for disjoint subsets $X, Y, Z$ of $U$, $I^S(X, Z, Y)$ if and only if $I(X, Z, Y)$ and $I(Y, Z, X)$.

### Proposition 4

Suppose $\rho$ is a GCPP over set of variables $U$.

(i) If $\rho(\top|x) = \rho(\top)$ for all $X \subseteq U$ and $x \in \underline{X}$ such that $\rho(x) \neq 0$, then $(I'_\rho)^R$ satisfies Trivial Independence.

(ii) If for all disjoint $W, Y, X \subseteq U$ there exists a function $M$ such that for all $x \in \underline{X}$, $M(\rho_x^{WY}) = \rho_x^Y$ then $(I'_\rho)^R$ satisfies Decomposition.

(iii) If for all disjoint $W, Y, X \subseteq U$ there exists a function $C$ such that for all $x \in \underline{X}$, $C(\rho_x^{WY}) = \rho_x^{W|Y}$ then $(I'_\rho)^R$ satisfies Weak Union.

(iv) If $\rho$ is a full GCPP and for all disjoint $W, Y, X \subseteq U$ there exists a function $J$ such that (a) for all $x \in \underline{X}$, $J(\rho_x^{W|Y}, \rho_x^Y) = \rho_x^{WY}$ and (b) $J(g, h) = J(g', h)$ when for all $w$ and $y$, $[g(w|y) = g'(w|y)$ or $h(y) = 0]$, for functions $g, g': W|Y \to D$ and $h: Y \to D$ with $(g, h)$ and $(g', h)$ in the domain of $J$. Then $(I'_\rho)^R$ satisfies Contraction.

(v) If $\rho$ is non-zero on $\underline{U}$ and for all disjoint $W, Y, X \subseteq U$ there exists a function $S$ such that for all $x \in \underline{X}$, $S(\rho_x^{W|Y}, \rho_x^{Y|W}) = \rho_x^{WY}$ then $(I'_\rho)^R$ satisfies Intersection.

The function $M$ in (ii) can be thought of as a marginalisation operator, and $C$ in (iii) as a conditioning operator. $J$ in (iv) gives a way of calculating the joint distribution from the conditional and marginal distributions; condition (b) in (iv) can be omitted if $\rho$ is non-zero on $\underline{U}$; $J$ is essentially just pointwise multiplication for probability and Dempster possibility.

The existence of $M$ in (ii) means that for each $x$, the joint distribution of $\rho_x$ on $WY$ determines the marginal distribution (of $\rho_x$ on $Y$). To see how this condition is used, suppose $I'_\rho(WY, \emptyset, X)$ and $\rho(x), \rho(x') \neq 0$; then $\rho_x^{WY} = \rho_{x'}^{WY}$ so $\rho_x^Y = \rho_{x'}^Y$ which leads to $I'_\rho(Y, \emptyset, X)$. Similar considerations apply to (iii), (iv) and (v).

Proposition 4 implies that if $\rho$ is a full GCPP, satisfying the conditions of (i), (ii), (iii) and (iv) above, and $I'_\rho$ is symmetric then $I'_\rho$ is a semi-graphoid; furthermore if $U$ is $\rho$-connected then $I'_\rho$ is a graphoid.

The result also leads to a way of constructing a semi-graphoid from a GCPP even if $I_\rho$ and $I'_\rho$ are not symmetric.

**Proposition 5**

If GCPP $\rho$ over $U$ is weakly conditional-coherent, and satisfies the conditions of (i), (ii), (iii) and (iv) of proposition 4 then $I_\rho^S$ is a semi-graphoid. If, in addition, $\rho$ satisfies the conditions of (v) then $I_\rho^S$ is a graphoid.

## 6   QUALITATIVE CONDITIONAL PROBABILITY

Another way of viewing GCPPs is as Qualitative Conditional Probabilities on Product spaces (QCPPs).

A Symmetric QCPP (abbreviated to SQCPP) $\approx$ over set of variables $U$ is an equivalence relation on $U^* \cup \{0,\infty\}$ satisfying, for disjoint $X, Y \subseteq U$ and $x \in \underline{X}$,

(i) $\rho(x) \approx 0$ if and only if for all $y \in \underline{Y}$, $\rho(xy) \approx 0$, and

(ii) for any $y \in \underline{Y}$, $\rho(x|y) \approx \infty$ if and only if $\rho(y) \approx 0$.

$\approx$ is said to be consistent if $0 \not\approx \infty$.

Independence structures $I_\approx$ and $I'_\approx$ on $U$ are defined analogously to the definitions for GCPPs: $I_\approx(X, Z, Y)$ $\iff$ $x|yz \approx x|y$ for all $x, y, z$ such that $yz \not\approx 0$, and $I'_\approx(X, Z, Y) \iff x|yz \approx x|y'z$ for all $x, y, y', z$ such that $yz \not\approx 0 \not\approx y'z$.

The framework of consistent SQCPPs is essentially equivalent to the framework of GCPPs. For any GCPP $\rho$ over $U$ we can define a consistent SQCPP $\approx_\rho$ over $U$, by first extending $\rho$ to $U^* \cup \{0, \infty\}$ by defining $\rho(0) = 0$ and $\rho(\infty) = \infty$, and then, for $\phi, \psi \in U^* \cup \{0, \infty\}$, defining $\psi \approx_\rho \phi \iff \rho(\psi) = \rho(\phi)$. We then have $I_{\approx_\rho} = I_\rho$ and $I'_{\approx_\rho} = I'_\rho$.

Conversely, for consistent SQCPP $\approx$ we can define GCPP $\rho_\approx$ by letting $D = (U^* \cup \{0, \infty\})/\approx$, calling the equivalence classes of 0 and $\infty$ by 0 and $\infty$ respectively and, for $x|y \in U^*$, defining $\rho_\approx(x|y) = d \iff d \ni x|y$. We have $I_{\rho_\approx} = I_\approx$ and $I'_{\rho_\approx} = I'_\approx$. Also, for any SQCPP $\approx$, we have $\approx_{(\rho_\approx)} = \approx$.

Relation $\preceq$ is said to be a QCPP over set of variables $U$ if it is a reflexive transitive relation on $U^* \cup \{0, \infty\}$ such that its symmetric part $\approx$ (the intersection of $\preceq$ and $\succeq$) is a SQCPP over $U$.

QCPPs might be thought of as probability functions without the numbers; a statement such as $x|z \preceq y|z$ could mean 'given $z$, value $y$ is at least as probable as $x$'.

QCPPs generate independence structures via their associated SQCPPs. The correspondences between GCPPs and SQCPPs mean that the sufficient conditions for independence structures to satisfy the graphoid properties given in section 5 can be translated into sufficient conditions for SQCPPs (and hence QCPPs) to generate independence structures with those properties.

If consistent $\approx$ is a full SQCPP (i.e, for all $x|y \in U^*$, $x|y \approx 0 \iff xy \approx 0$ and $y \not\approx 0$), a sufficient condition for $I_\approx$ to satisfy Symmetry is the following 'cross multiplication' condition:

For all disjoint $X, Y, Z \subseteq U$, $x \in \underline{X}, y \in \underline{Y}, z \in \underline{Z}$, if $xz \not\approx 0$ and $xyz|yz \approx xz|z$ then $xyz|xz \approx yz|z$,

where we have trivially extended $\approx$ to elements $xy|y$ of $XY|Y$ (for disjoint $X, Y \subseteq U$), by placing $xy|y$ in the same $\approx$-equivalence class as $x|y$.

### Constructing QCPPs

We will often want to construct a QCPP from a number of different types of information:

(i) some qualitative probability relationships we expect always to hold, such as, perhaps, $0 \preceq x|y$ for all $x|y \in U^*$;

(ii) some desirable properties of $\preceq$, such as the above sufficient condition for Symmetry of $I_\approx$, and other conditions that imply graphoid properties;

(iii) an agent's comparative probability judgements, e.g., statements of the form $x|z \preceq x$ or $x|z \approx y|z$;

(iv) an agent's conditional independence judgements.

The obvious way to attempting to construct a QCPP for a particular situation is to treat (i) and (iii) as sets of axioms and (ii) and (iv) as sets of inference rules, and generate the QCPP from these. However, there is a technical problem: because of the conditions $yz \not\approx 0$, the conditional independence assumptions cannot quite be viewed as sets of inference rules. We can solve this by requiring that the user gives (explicitly or implicitly) *all* the values $u$ of $U$ such that $u \approx 0$ (that is, the set of all $u$ which are considered impossible); the key point here is that the application of the rules must not lead to any more zero values of $U$. The conditional independence assumptions can now be viewed as inference rules, since they are now closed under intersection. For the same reason, we require the properties in (ii) also to be closed under intersection, once the zeros of $U$ are determined.

Naturally, if we have included in (ii) properties which imply that $I_\approx$ is a semi-graphoid, then we can propagate conditional independence assumptions using the semi-graphoid properties, or using the graphical methods described in [Pearl, 88].

## 7  COMPUTATION OF GCPPS

Here we only consider computation of values of the joint distribution of a GCPP, leaving other aspects of computation for future work.

Let $I$ be an independence structure on $U$. For $X_i \in U$, and $W \subseteq U \setminus \{X_i\}$ the set $B \subseteq W$ is said to be an $I$–Markov boundary of $X_i$ with respect to $W$ if $B$ is minimal such that $I(\{X_i\}, B, W \setminus B)$.

If $I$ satisfies Trivial Independence then there is at least one $I$–Markov boundary of $X_i$ with respect to $W$, and if $I$ satisfies Weak Union and Intersection then there is at most one.

### Proposition 6

Let $\rho$ be a GCPP over $U$, which is labelled $X_1, \ldots, X_n$. Suppose there exists a function[6] $\diamond\colon D \times D \to D$ such that for all $x|y \in U^*$, $\rho(xy) = \rho(x|y) \diamond \rho(y)$. Then for any $x_i \in \underline{X}_i$, $i = 1, \ldots, n$,

$$\rho(x_1 \cdots x_n) = \rho(x_n|b_n) \diamond \cdots \diamond \rho(x_1|b_1),$$

where the repeated application of $\diamond$ is performed right-to-left, $b_i$ is $x_1 \cdots x_n$ projected onto $B_i$, and $B_i$ is an $I_\rho$–Markov boundary of $X_i$ with respect to $\{X_1, \ldots, X_{i-1}\}$.

The Boundary Directed Acyclic Graph [Pearl, 88] is formed by letting the parents of each $X_i$ be $B_i$. The above result shows that, just like for probability, the values of the joint distribution (i.e, $\rho$ on $\underline{U}$) can be calculated using this DAG together with, for each $X_i$, the matrix giving the values of $\rho$ on $\underline{X}_i$ conditional on the values of the parents of $X_i$.

If $I_\rho$ is a semi-graphoid, then a result in [Verma, 86] (also Theorem 9 of [Pearl, 88]) implies that the Boundary DAG is a minimal I-map for $I_\rho$ so that conditional independence properties of $I_\rho$ can be derived by testing for d-separation in the DAG.

## 8  DISCUSSION

The sufficient conditions we have found for GCPPs to generate semi-graphoids seem natural and fairly weak. However, we clearly we need to look for more (sensible) examples of GCPPs that generate semi-graphoids, via our sufficient conditions. For example, it would be desirable to find simple such uncertainty formalisms which take values which are not totally ordered.

For Qualitative Conditional Probability, we need to consider appropriate extra axioms and inference rules to add to the system. The relationship between Qualitative Conditional Probability and relations on conditional objects [Goodman et al, 91] would be interesting to explore, as would its relationship with comparative probability [Walley and Fine, 79]. There may well also be connections between GCPPs and the framework of [Shenoy, 92] which uses a product definition of independence.

### Acknowledgements

Thanks to Serafín Moral and Luis de Campos for some useful discussions, and to Milan Studený and the referees for their helpful comments. The author is supported by a SERC postdoctoral fellowship. I am also grateful for the use of the computing facilities of the school of Computing and Mathematical Sciences, Oxford Brookes University.

---

[6]The existence of this function is very closely related to one of the axioms in a justification of Bayesian probability [Cox, 46].